\documentclass[sigconf,nonacm]{acmart}
\usepackage{graphicx}
\usepackage{comment}
\usepackage{amsmath}
\usepackage{color}
\usepackage{subcaption}
\usepackage{booktabs}
\usepackage{multirow}
\usepackage{array}

\definecolor{corner}{RGB}{33,102,172}
\definecolor{sidewalk}{RGB}{239,138,98}
\definecolor{crossing}{RGB}{178,24,43}
\definecolor{classtext}{RGB}{253,219,199}

\definecolor{Red}{RGB}{225, 25, 5}
\definecolor{Green}{RGB}{51, 160, 44}
\definecolor{Blue}{RGB}{0, 100 ,200}
\definecolor{Yellow}{RGB}{255, 255 ,0}

\definecolor{BrightRed}{RGB}{225, 0, 0}
\definecolor{LightYellow}{RGB}{254, 254 ,139}

\AtBeginDocument{%
  \providecommand\BibTeX{{%
    \normalfont B\kern-0.5em{\scshape i\kern-0.25em b}\kern-0.8em\TeX}}}

\makeatletter
\def\@copyrightspace{\relax}
\makeatother

\begin{document}

\title{APE: An Open and Shared Annotated Dataset for Learning Urban Pedestrian Path Networks}

\author{Yuxiang~Zhang,
        Nicholas-J~Bolten,
        Sachin~Mehta,
        and~Anat~Caspi 
}
\authornote{Preprint. Under review.}
\authornote{Corresponding author: Yuxiang Zhang (email: yz325@uw.edu)}
\affiliation{
  \institution{ University of Washington}
  \city{Seattle}
  \state{Washington}
  \country{USA}
  }

\renewcommand{\shortauthors}{Zhang et al.}

\begin{abstract}
Inferring the full transportation network, including sidewalks and cycleways, is crucial for many automated systems, including autonomous driving, multi-modal navigation, trip planning, mobility simulations, and freight management. Many transportation decisions can be informed based on an accurate pedestrian network, its interactions, and connectivity with the road networks of other modes of travel. A connected pedestrian path network is vital to transportation activities, as sidewalks and crossings connect pedestrians to other modes of transportation. However, information about these paths' location and connectivity is often missing or inaccurate in city planning systems and wayfinding applications, causing severe information gaps and errors for planners and pedestrians. This work begins to address this problem at scale by introducing a novel dataset of aerial satellite imagery, street map imagery, and rasterized annotations of sidewalks, crossings, and corner bulbs in urban cities. The dataset spans $2,700 km^2$ land area, covering select regions from $6$ different cities. It can be used for various learning tasks related to segmenting and understanding pedestrian environments. We also present an end-to-end process for inferring a connected pedestrian path network map using street network information and our proposed dataset. The process features the use of a multi-input segmentation network trained on our dataset to predict important classes in the pedestrian environment and then generate a connected pedestrian path network. Our results demonstrate that the dataset is sufficiently large to train common segmentation models yielding accurate, robust pedestrian path networks.
\end{abstract}

%    <concept>
%        <concept_id>10010147.10010257</concept_id>
%        <concept_desc>Computing methodologies~Machine learning</concept_desc>
%        <concept_significance>500</concept_significance>
%        </concept>
%    <concept>
%        <concept_id>10010147.10010178.10010224</concept_id>
%        <concept_desc>Computing methodologies~Computer vision</concept_desc>
%        <concept_significance>500</concept_significance>
%        </concept>
%    <concept>
%        <concept_id>10010405.10010481.10010485</concept_id>
%        <concept_desc>Applied computing~Transportation</concept_desc>
%        <concept_significance>500</concept_significance>
%        </concept>
%    <concept>
%        <concept_id>10003120.10011738.10011776</concept_id>
%        <concept_desc>Human-centered computing~Accessibility systems and tools</concept_desc>
%        <concept_significance>500</concept_significance>
%        </concept>
%  </ccs2012>

% \ccsdesc[500]{Computing methodologies~Machine learning}
% \ccsdesc[500]{Computing methodologies~Computer vision}
% \ccsdesc[500]{Applied computing~Transportation}
% \ccsdesc[500]{Human-centered computing~Accessibility systems and tools}

\keywords{dataset, accessibility, computer vision, pedestrian, urban network}


\maketitle

\section{Introduction}
\label{sec:intro}

A connected pedestrian path network detailing the location and connectivity of sidewalks, crossings, and curbs is essential to any transportation network. Sidewalks and crossings connect all other transportation modes and are the key to building an accessible city. Creating a fully connected pedestrian path network is essential for many city planning tasks and wayfinding applications \cite{bolten2022towards}.

While automobile roads have been extensively mapped \cite{mattyus2017deeproadmapper, mi2021hdmapgen}, mapping information for the paths that serve pedestrians is incomplete or missing. This fundamental lack of information creates bottlenecks in the transportation network and artificially disconnects neighborhoods. More importantly, this information gap impacts pedestrians with disabilities' travel in the cities. A connected pedestrian path network map consisting of the location and connectivity of sidewalks and crossings would close this information gap and improve accessibility for all.
% *** Ricky: I am not sure the meaning of travel-underserved in the last part of the sentence, removing it for now ***
 % A connected pedestrian path network map consisting of the location and connectivity of sidewalks and crossings will alleviate this information gap and promote accessibility for all people, including travel- \emph{underserved} communities, to navigate along sidewalks. 

\begin{figure}[htbp] 
    \centering
    \begin{subfigure}[]{0.49\columnwidth}
        \centering
        \includegraphics[width=\columnwidth]{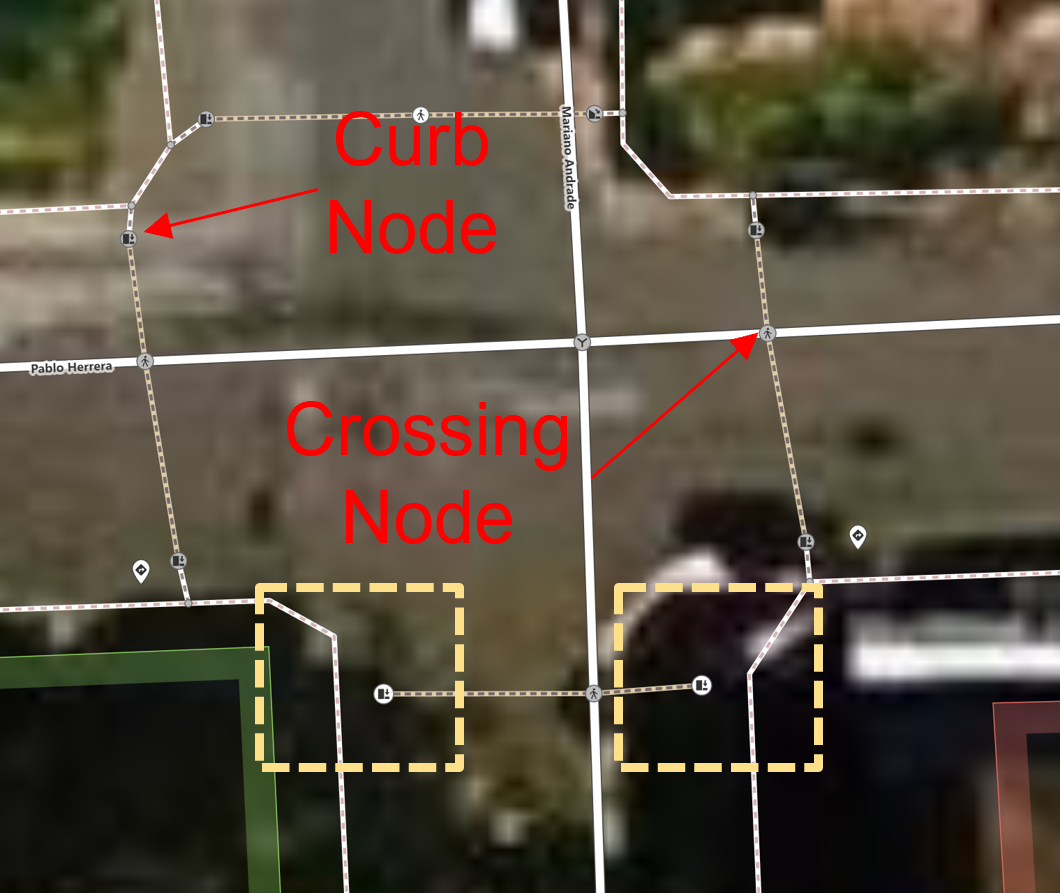}
        \caption{}
        \label{fig:annotation_error_connect}
    \end{subfigure}
    \begin{subfigure}[]{0.49\columnwidth}
        \centering
        \includegraphics[width=\columnwidth]{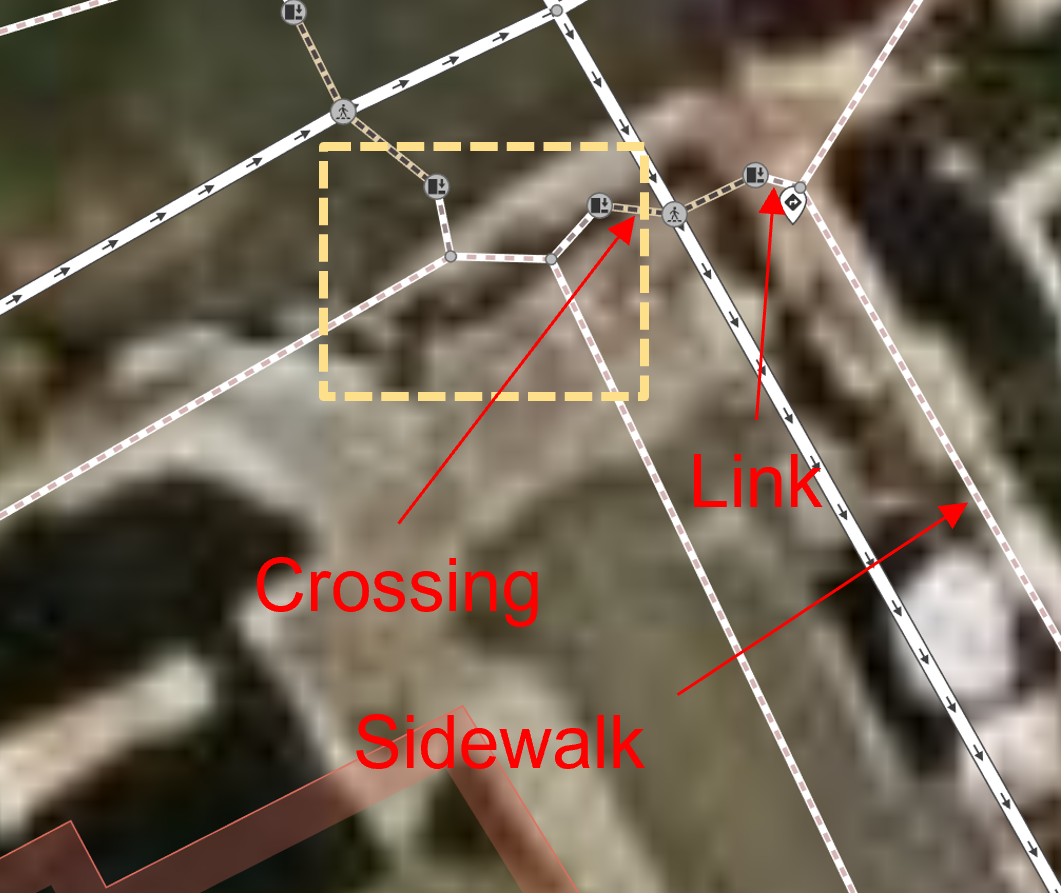}
        \caption{}
        \label{fig:annotation_error_loc}
    \end{subfigure}
    \caption{Example of annotations from human mappers. \colorbox{BrightRed}{Red text} shows different types of features in a pedestrian path network map. \colorbox{LightYellow}{Yellow} boxes show examples of errors made by human mapper:  (a) crossing lines are not connected to the sidewalk lines with link lines (b) the curb nodes are falsely mapped to the middle of the road.}
    \label{fig:annotation_error}
\end{figure}

% this paragraph talks about the challenge
A pedestrian map requires different information than a map for car navigation. It must show common paths (sidewalks, crossings), their connectivity, transitions, and additional attributes of those paths. Altogether this is difficult to gather manually due to the large amount of data and potential for error \cite{bolten2021towards, bolten2022towards}. In addition, mappings that are done by human mappers often contain errors due to many factors, including low image quality and ambiguous mapping protocols. We validated and evaluated several areas mapped through a human annotation campaign carried out by the OpenSidewalks Project \cite{opensidewalks}. We found that although mappers were trained before doing the mapping and were provided with instructions and mapping protocol, different types of errors were still made in the mapping process. Example errors are visualized in Figure \ref{fig:annotation_error}, and Table \ref{tab:error_cnt} summarizes our analysis on human-generated maps, where we analyzed the three elemental geometry features ( sidewalks,  crossings, and sidewalk links that connect sidewalks to crossings) \footnote{The nodes that are analyzed in Table 1b include the curb nodes (the intermediate nodes shared by a crossing line and a link line that connects a crossing to a sidewalk), and the crossing nodes (the nodes at the intersection of a crossing line and a road line). Sidewalks are mapped as lines in common GIS data thus there is no designated tag for sidewalk nodes in the mapped annotations.}. These geometry-based errors can lead to inaccurate depictions of pedestrian paths and internal path connectivity, hindering downstream applications and particularly impacting routing algorithms and network analysis. Machine learning methods that globally generate connected pedestrian path networks efficiently would improve mapping accuracy, reduce manual efforts, and improve transportation planning.

\begin{table}[htbp]
\centering
\caption{Analysis on human annotations}
\begin{subtable}[]{1\columnwidth}
    \centering
    \caption{Statistics of validated annotations}
    \label{tab:oswv}
    \begin{tabular}{l|cc}
    \toprule
        {}  &  \textbf{Annotation count } & \textbf{Annotation length} \\
        \midrule
        links  &  1526 & 9423.9 m \\
        sidewalks  &  3537 & 37527.1 m \\
        crossings  &  1594 & 8543.9 m\\
    \bottomrule
\end{tabular}
\end{subtable}
\newline
\vspace*{0.2 cm}
\newline
\begin{subtable}[]{1\columnwidth}
    \centering
    \caption{Number of errors on Nodes}
    \begin{tabular}{l|ccc}
    \toprule
       \textbf{Node type}  &  \textbf{Location } & \textbf{Classification} & \textbf{Tag} \\
        \midrule
       curb  &  301 & 108 & 1\\
       crossing  &  51 & 35 & 226 \\

    \bottomrule
\end{tabular}
\label{tab:error_cnt_node}
\end{subtable}
\newline
\vspace*{0.2 cm}
\newline
\begin{subtable}[]{1\columnwidth}
    \centering
    \caption{Number of errors on LineStrings}
    \begin{tabular}{l|ccc}
    \toprule
       \textbf{Line type}  &  \textbf{Location } & \textbf{Classification} & \textbf{Tag} \\
        \midrule
       links  & 337 & N/A & 59\\
       sidewalks  & 396 & N/A & 59\\
       crossing  & 375 & 15 & 0 \\
    \bottomrule
\end{tabular}
\label{tab:error_cnt_line}
\end{subtable}
\label{tab:error_cnt}
\end{table}

%\subsection{Contributions} 
In this work, we present a novel dataset, the Annotations for Pedestrian Environment (APE) dataset, created through a rasterization process of GIS mapping annotations. The APE dataset includes aerial satellite images, street map tiles, and rasterized annotations. The contributions of this work are two folds (1) we present a method for rasterizing Geographic Information System (GIS) mapping annotations to generate the APE dataset (Figure \ref{fig:data_sample}). (2) We also develop an end-to-end process (Figure \ref{fig:flow_chart}) to infer a connected pedestrian path network, using a multi-input segmentation network trained on the APE dataset to predict pedestrian path locations and integrate them with street network data to complete a pedestrian path network. The APE dataset and code will be released upon publication.

The rest of the paper is organized as follows. Section \ref{sec:related} discusses related work. Section \ref{sec:dataset} introduces the APE dataset. Section \ref{sec:method} details the implementation of our process. Experiment results are shown in Section \ref{sec:exp}. Discussions and conclusions are made in Section \ref{sec:dis}.

\section{Related Work}
\label{sec:related}
We discuss related work in three domains: Section \ref{sec:related_street}: studies on street/road network mapping, Section \ref{sec:related_pedes}: methods to map pedestrian environments, Section \ref{sec:related_data}: available aerial imagery datasets. The combined work highlights the need for an imagery-based dataset that targets the pedestrian environment.   

\subsection{Street and Road Network Map Generation}
\label{sec:related_street}
Studies have investigated using aerial imagery along with auxiliary data for street mapping. Wu et al.  \cite{wu2019road} used OpenStreetMap (OSM) centerlines as labeled data and extracted roads from very high-resolution (VHR) satellite images. Sun et al. \cite{sun2019leveraging} added crowd-sourced global positioning system (GPS) data to satellite images to extract roads with CNN-based semantic segmentation. Zhou et al. \cite{zhou2021funet} fused remote sensing images and GPS for road detection and extraction. Additional recent learning-based studies included Lu et al.  \cite{lu2021gamsnet} proposing a  multi-scale residual network for road detection, Pan et al. \cite{pan2021generic} proposing a fully convolutional neural network using VHR remote sensing, Mattyus et al. \cite{mattyus2017deeproadmapper} estimated road topology from aerial images, Mi et al. \cite{mi2021hdmapgen} generated road lane graphs from LiDAR data with a hierarchical graph generation model. Importantly, work in this domain solely focuses on automobile road detection and extraction, and does not address the generalization or extension of the proposed methods to the pedestrian environment.  Methods for mapping the environments that serve pedestrians' travel have not been widely studied. 

\subsection{Pedestrian Environment Mapping}
\label{sec:related_pedes}
Few studies have focused on mapping pedestrian environments compared to automobile roads. Karimi et al. \cite{karimi2013pedestrian} explored pedestrian map generation approaches in a small-scale area, demonstrating preliminary mapping results that heavily depended on the availability and quality of input data.  Recent advancements in remote sensing have led to more imagery-based approaches, such as Ahmetovic et al. \cite{ahmetovic2015zebra} detection of zebra crossings using satellite imagery and validation with street-level images. Likewise, Ghilardi et al. \cite{ghilardi2016crosswalk} classified and located crosswalks using an SVM classifier over data extracted from road maps, and Ning et al. \cite{ning2022sidewalk} extracted sidewalks from aerial images with a neural network and restored occluded segments from street view images. These studies improved pedestrian environment mapping, but they do not generate a comprehensive, connected pedestrian path network needed for city planning and navigation. Other studies use on-the-ground data, such as Zhang et al. \cite{zhang2021collecting} automated collection of street-view images with auxiliary data to map sidewalk connectivity and sidewalk infrastructure., Hou et al. \cite{hou2020network} extraction of sidewalks using LiDAR data and point cloud segmentation. However, these methods often require physical systems to cover a large area to generate a pedestrian path network. Other studies that map pedestrian path networks at scale are often based solely on existing street (road) data. For example, Li et al. \cite{li2018semi}'s semi-automated method generated a sidewalk network using parcel-level data and roadway centerline, but it required human editing for quality control.

\subsection{Aerial Imagery Datasets}
\label{sec:related_data}
There are several datasets containing aerial imagery for mapping. The TorontoCity dataset \cite{wang2016torontocity} contains aerial satellite images for road curb extraction and road centerline estimation. PRRS \cite{aksoy2008performance} presents a dataset for building extraction and Digital Surface Model (DSM) estimation using satellite data. In addition, the DeepGlobe dataset \cite{DeepGlobe18} and the ISPRS dataset \cite{ISPRS} contain imagery and annotations for tasks including road extraction, building detection, and land cover classification. These datasets enable researchers to study different tasks involving the use of aerial imagery data, but they neither directly nor specifically target the pedestrian path networks, nor provide annotations needed for semantically understanding the pedestrian environment. In sum, currently, there are no extensive datasets focused on pedestrian environments, nor methods to generate pedestrian path networks for large areas.

\section{the Annotations for Pedestrian Environment dataset}
\label{sec:dataset}

\begin{figure}[htbp] 
    \centering
    \resizebox{\columnwidth}{!}{
        \begin{tabular}{ccc}
             \textbf{Aerial satellite image} & \textbf{Street map image} & \textbf{Rasterized annotation} \\
             \includegraphics[width=0.33\columnwidth]{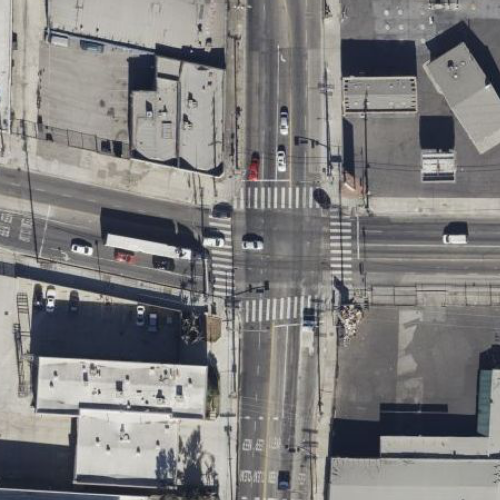} & 
             \includegraphics[width=0.33\columnwidth]{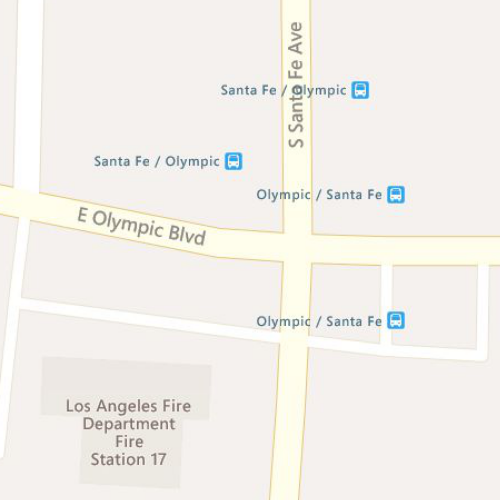} & 
             \includegraphics[width=0.33\columnwidth]{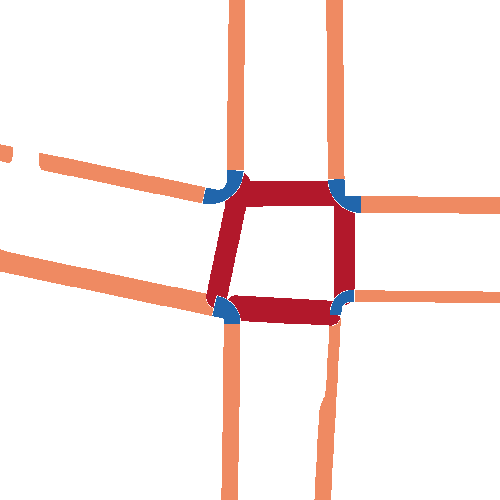} \\
             \includegraphics[width=0.33\columnwidth]{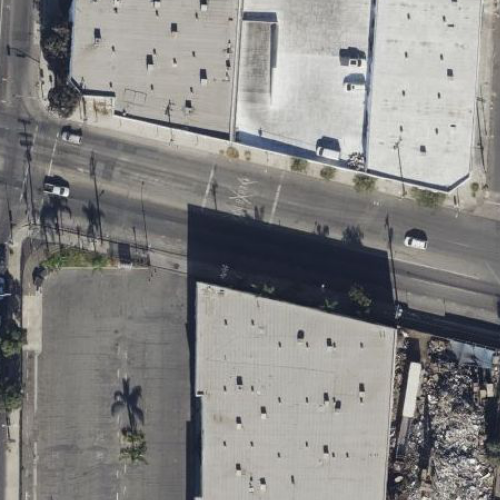} & 
             \includegraphics[width=0.33\columnwidth]{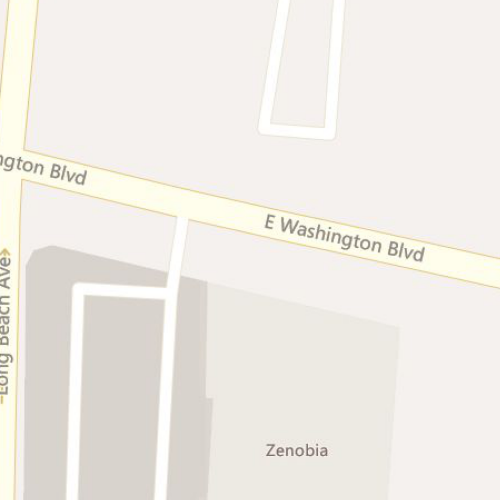} & 
             \includegraphics[width=0.33\columnwidth]{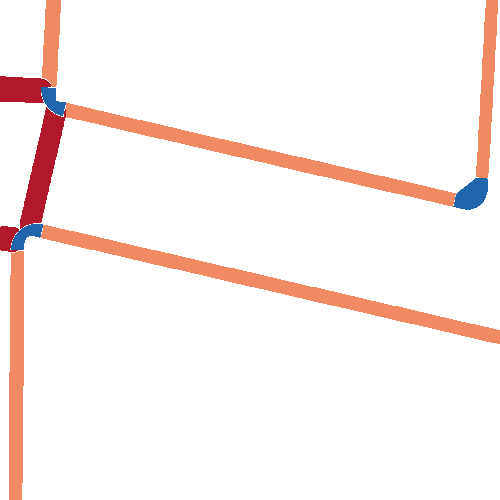}\\
              \includegraphics[width=0.33\columnwidth]{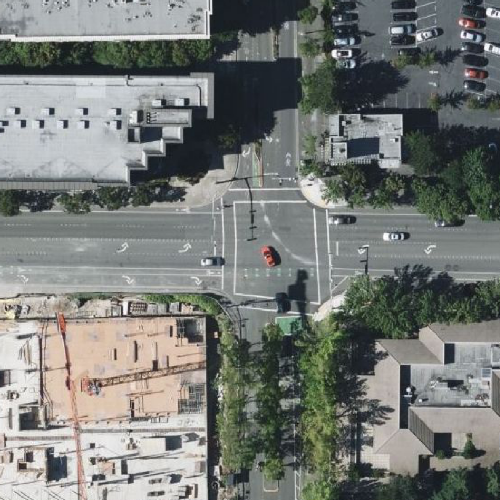} & \includegraphics[width=0.33\columnwidth]{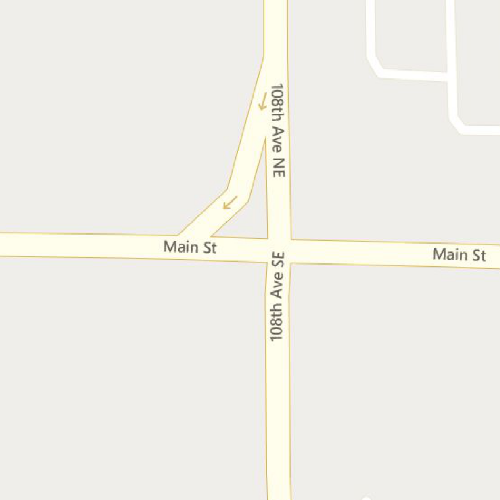} & \includegraphics[width=0.33\columnwidth]{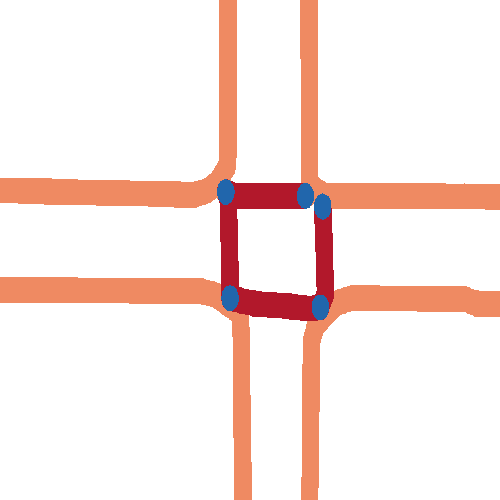} \\
               \includegraphics[width=0.33\columnwidth]{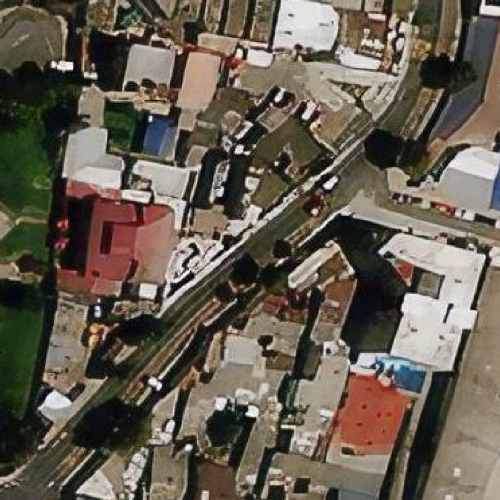} & \includegraphics[width=0.33\columnwidth]{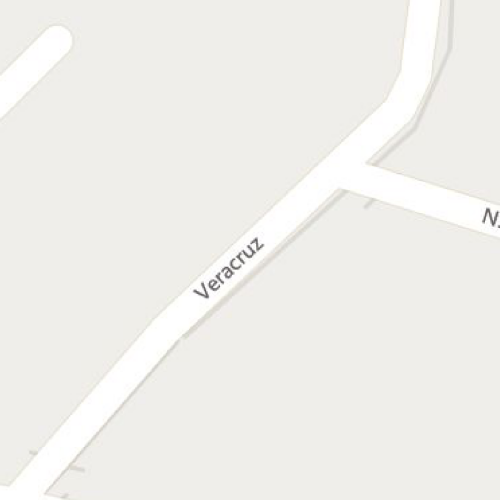} & \includegraphics[width=0.33\columnwidth]{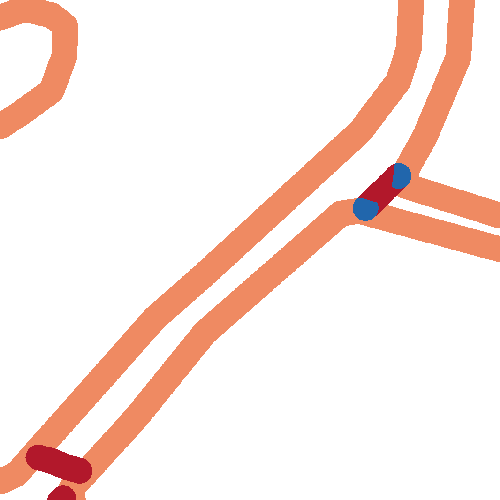} \\
              \includegraphics[width=0.33\columnwidth]{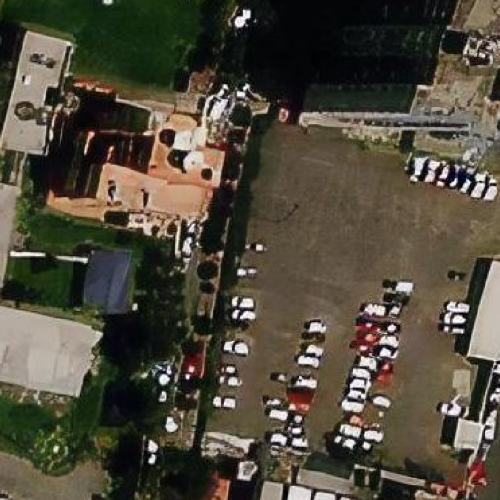} & \includegraphics[width=0.33\columnwidth]{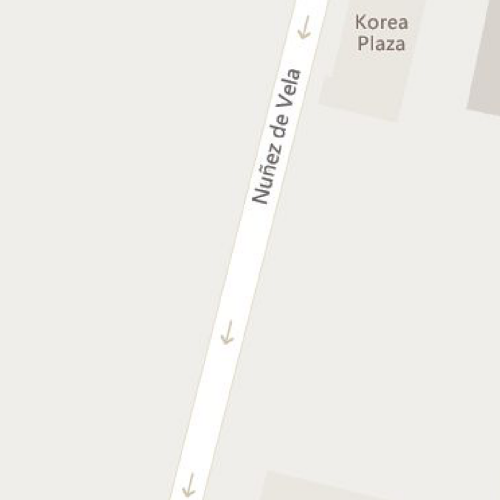} & \includegraphics[width=0.33\columnwidth]{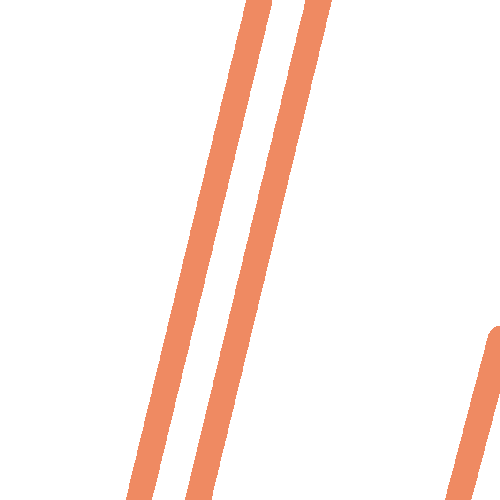} \\
        \end{tabular}
    }
    \caption{Samples from the APE dataset. The first three rows show samples from North American cities. The last two rows show samples from South American cities, where the available aerial satellite images have lower resolutions.}
    \label{fig:data_sample}
\end{figure}

To address the lack of large-scale datasets that specifically target the pedestrian environment, we develop a method to rasterize more commonly found GIS annotations and present the Annotations for Pedestrian Environment (APE) dataset, which contains labels for essential classes that compose a connected pedestrian path network graph.
The GIS annotations for pedestrian path environments, either created by city agencies or crowdsourcing by local mappers (e.g. OpenStreetMap (OSM)), are primarily created for use in vector GIS data systems. The commonly used GIS data formats (e.g. Shapefile map or GeoJSON) cannot be directly used in computer vision tasks that required input data to be imagery-like. In creating the APE dataset, we rasterize GIS data from multiple sources in multiple geographic areas to create the labeled data needed for understanding pedestrian environments in computer vision tasks.

\subsection{Coverage}
\label{sec:coverage}
The APE dataset covers select regions in the following areas: (1) Los Angeles, CA, United States (2) Bellevue, WA, United States (3) Quito, Ecuador (4) Sao Paulo, Brazil (5) Santiago, Chile (6) Gran Valparaiso, Chile. Consequently, the APE dataset encompasses diverse built environments from both North and South America. Cities in South America are more densely populated with higher street and intersection density, but simpler in shape compared to North American cities\cite{sarmiento2021built}. The OpenSidewalks Project \cite{opensidewalks} provides crowdsourced GIS annotations for select regions in each of these 6 cities in a consistent schema \cite{opensidewalksSchema}, with annotations created by trained and veteran mappers. \footnote{A mapper is considered trained if the mapper is given clear written instructions and had completed at least one mapping training session. A trained mapper is considered to be a veteran mapper if the mapper has completed several mapping tasks and the mapping outcomes are validated in a peer-review process.} These annotations are available in standard GIS formats in OSM. In addition, the City of Los Angeles provides a GIS dataset with additional information for the Los Angeles area, including annotations for sidewalks, crossings, corner bulbs, and driveways \cite{LAHub}. The APE dataset consists of a total of $14,800$ samples and spans approximately $2,700 km^2$ land area. 

\subsection{Collecting Images and Annotations}
\label{sec:collection}
Each dataset sample includes (1) the aerial satellite image, (2) the street map imagery tile, and (3) the rasterized annotations for pedestrian paths. The aerial satellite images and the street map imagery tiles are acquired from Bing Maps along every major road for each area described in Section \ref{sec:coverage}. 
The GIS annotations mainly come in three geometry types: \textit{Point}, \textit{LineString}, and \textit{Polygon}. The nodes representing crossings, curbs, and link endpoints are usually annotated as the \textit{Point} geometry. For each of these annotated objects with the \textit{Point} geometry, we convert the point to a circle by adding a buffer zone with a fixed radius and then rasterize the circle in the image. Sidewalks and crossings usually have two types of representations: the centerline representation and the polygon boundary representation. (1) If they are represented by their centerline as the \textit{LineString} geometry, for each \textit{LineString}, we convert it into a \textit{Polygon} by adding a buffer zone to each side of the \textit{LineString}, then rasterize the \textit{Polygon} as a filled polygon shape in the image. (2) If they are already annotated by their boundary as the \textit{Polygon} geometry, we directly rasterize them as filled polygon shapes in the image. \footnote{For the data currently in the APE dataset, the polygon boundary representation is only applied to the sidewalk annotations from the City of Los Angeles. The sidewalk annotations from the OpenSidewalks project are all in the centerline representation per the OpenSidewalks standard.} Each set of an aerial satellite image, a street map image, and a rasterized annotation are aligned with their precise geographic bounding boxes.

\subsection{Classes}
\label{sec:classes}
Our dataset provides annotations for three distinct classes needed to semantically segment the pedestrian environment. Shown in Figure \ref{fig:data_sample}, these classes include (1)  \colorbox{corner}{\textcolor{classtext}{\textit{Corner bulb}}}  (2) \colorbox{sidewalk}{\textcolor{classtext}{\textit{Sidewalk}}}  (3) \colorbox{crossing}{\textcolor{classtext}{\textit{Crossing}}}. Corner bulbs are commonly used when describing a transportation network since they serve as a transition zone connecting a sidewalk to curb ramps, crossings, or another sidewalk. The nodes representing sidewalk endpoints, link endpoints, and curbs are usually located within the corner bulbs. Sidewalks and crossings are essential elements in an urban pedestrian path network graph, as the lines representing sidewalks and crossings are essentially the edges a pedestrian will traverse. The focus of our work is on mapping pedestrian paths, thus, all other annotated classes (including roads, buildings, and trees) collectively comprise \textit{background class} in the following experiments. However, these classes can also be rasterized with our method (Section \ref{sec:collection}) and added to the APE dataset as additional class labels.

\subsection{Challenges}
\label{sec:data_challenge}
As previously noted in autonomous-driving datasets, imagery dataset bias can be introduced by non-representative geographic locations of images \cite{torralba2011unbiased, wilson2019predictive, jo2020lessons}. This bias generally refers to a systematic error that results from the training dataset only representing a limited geographic region or a limited type of environment. To expand APE's ability to generalize to other regions and environments, we created a more balanced and representative dataset that includes images from a diverse range of regions and environments, including imagery from both North and South American urban contexts. However, this introduces a different challenge having to do with geopolitical biases in satellite image collection: only lower-resolution aerial satellite images were openly available for South American cities. This presented challenges in model learning. Figure \ref{fig:data_sample} displays samples from the APE dataset, with the first three rows representing North American cities and the last two rows representing lower-resolution images in South American cities. Differences in model performance when predicting pedestrian networks in different cities are discussed in Section \ref{sec:exp}.

Another challenge for learning from aerial satellite images is that the important classes such as sidewalks and crossings are occluded in some satellite images\cite{ning2022sidewalk}. For example, in the second and third rows shown in Figure \ref{fig:data_sample}, part of the pixels that are labeled as \textit{sidewalk} are occluded by buildings, the shade from buildings, and vegetation in the aerial satellite images.  In these cases, learning from aerial satellite images alone is very challenging and street map imagery tiles can provide crucial auxiliary information. Section \ref{sec:exp} shows a quantitative analysis of the models trained with (1) only aerial satellite images (2) only street map imagery tiles, and (3) both aerial satellite images and street map imagery tiles.

\section{Inferring a connected pedestrian path network graph with APE }
\label{sec:method}
\begin{figure*}[t!]
\centering
\includegraphics[width=1.75\columnwidth]{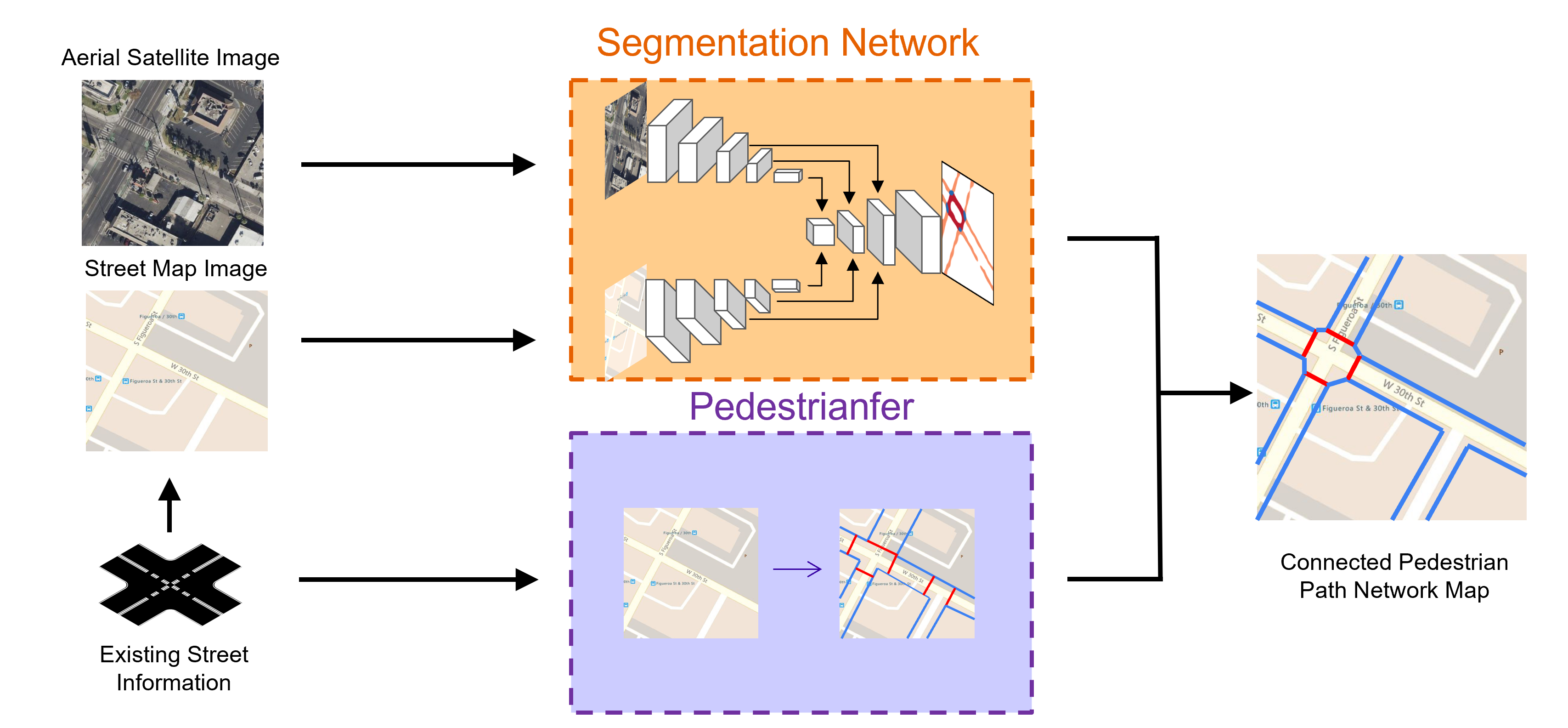}
\caption{Overview of the end-to-end process for inferring pedestrian path network graph: Section \ref{sec:pedestrianfer} outlines \textit{Pedestrianfer}, generating a pedestrian path network from incomplete information. Section \ref{sec:cnn} details the segmentation network using aerial and road map images to predict class labels of pixel locations in the pedestrian map. Section \ref{sec:opt} explains how \textit{Pedestrianfer} and segmentation network information are combined for accurate graph inference.}
\label{fig:flow_chart}
\end{figure*}

The APE dataset enables us to automate the process of generating pedestrian path network data at scale. In this section, we introduce an end-to-end process to infer a connected pedestrian path network. As shown in Figure \ref{fig:flow_chart}, the process consists of three main steps. First, Section \ref{sec:pedestrianfer} details the creation of a hypothesized graph with existing street network data using a tool we developed called \textit{Pedestrianfer}. Second, Section \ref{sec:cnn} describes the multi-input segmentation network trained on the APE dataset for generating pixel-wise prediction masks for the important classes in the pedestrian environment. Lastly, Section \ref{sec:opt} describes the process of using the predictions from the segmentation network to optimize the hypothesized graph into an accurate connected pedestrian path network graph.

\begin{figure}[htbp]
\centering
\includegraphics[width=0.7\columnwidth]{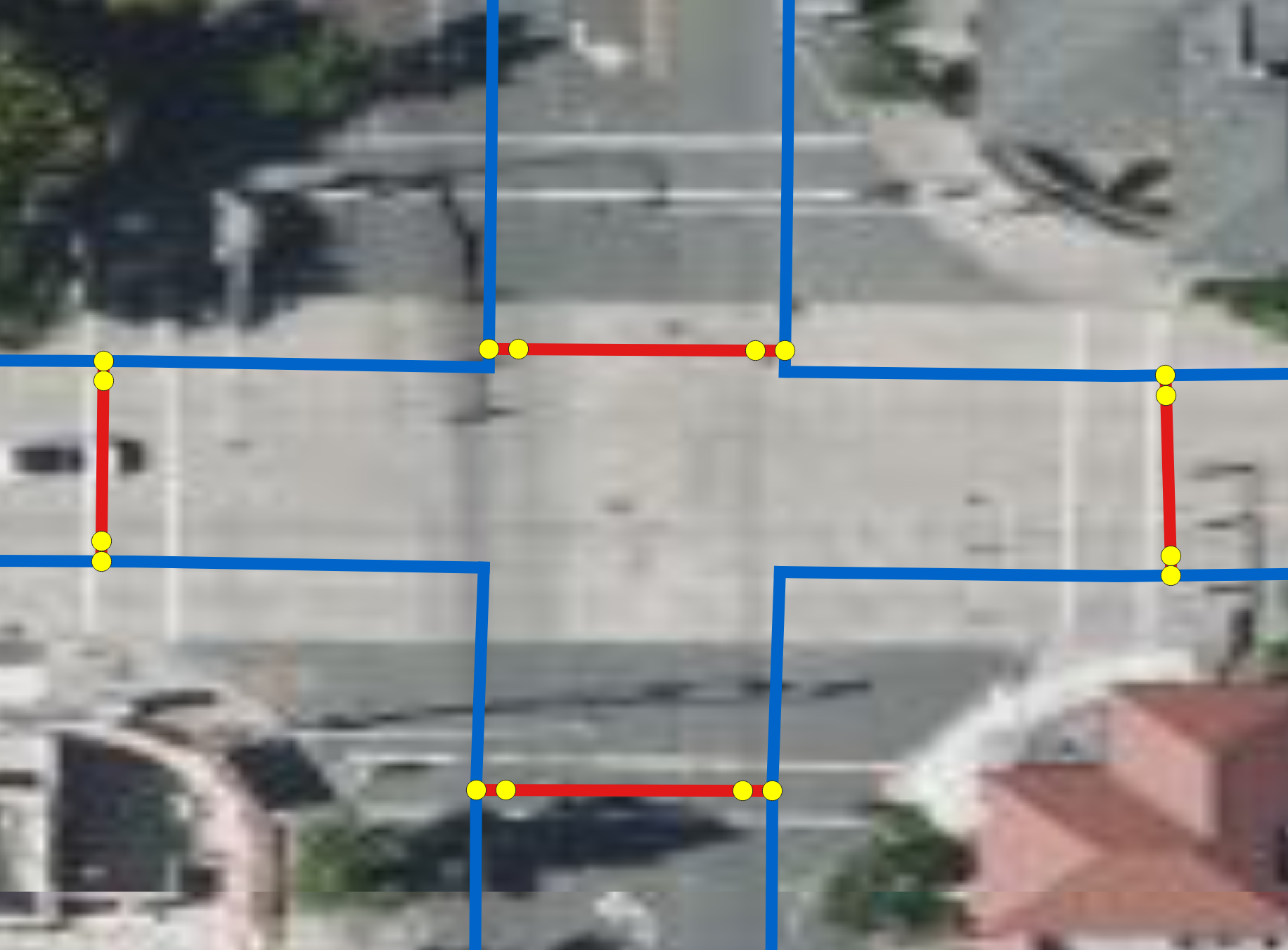}
\caption{Hypothesized graph generated with \textit{Pedestrianfer}. \colorbox{Yellow}{Yellow} dots represent the hypothesized nodes for the curb modes and the nodes on the sidewalk. \colorbox{Blue}{Blue} lines represent the hypothesized sidewalks. \colorbox{Red}{Red} lines represent the hypothesized crossings. Links are part of the crossings between the curb nodes and the nodes on the sidewalk}
\label{fig:hypo}
\end{figure}

\subsection{Pedestrianfer}
\label{sec:pedestrianfer}
To infer a hypothesized pedestrian path network from incomplete information, we developed \textit{Pedestrianfer} (Figure \ref{fig:hypo}). \textit{Pedestrianfer} performs three discrete inference tasks: (1) inferring an (optimistic) sidewalk network from the existing street networks, (2) inferring street crossing locations from a street network and a sidewalk network, and (3) creating a preliminary conjecture of curb interface locations around the street crossings. \textit{Pedestrainfer} is similar in outcome to the method \cite{li2018semi} discussed in Section \ref{sec:related} but with the following key differences: (1) the only required input to our method is a street network, and sidewalk locations are estimated from that street network, (2) full intersection-to-intersection sidewalk paths are estimated rather than being broken into 50-meter segments, and (3) street crossing pathways are generated via a cost function that weighs several properties and does not require manual intervention. Additionally, Pedestrianfer generates pathways according to the OpenSidewalks Schema specification \cite{opensidewalksSchema}.

\textit{Pedestrianfer} first infers sidewalk networks from a street network under two alternative regimes, each representing a different hypothesis on the built environment. The alternatives depend on whether the data source is a vector layer of street network data with metadata on the presence (and optionally, offset distance) of sidewalks. If metadata is present, sidewalks are placed only where indicated. If not, \textit{Pedestrianfer} retrieves a street network and hypothesizes full sidewalks, i.e. the scenario in which all streets have a connected sidewalk on either side. In the latter case, \textit{Pedestrianfer} retrieves an existing street network, such as a street network from OSM. In either case, the \textit{Pedestrianfer} process involves 4 steps: (1) creating a directed graph representation of the street network, (2) generating all right-hand-turn paths that start and stop at the same node (closed paths), (3) drawing sidewalks via line offset algorithm, and (4) trimming or joining them based on the path context, e.g. trimming overlapping sidewalk lines when they are neighbors along the path.

Next, \textit{Pedestrianfer} uses a street network and a sidewalk network (generated by \textit{Pedestrianfer} in the previous step if it does not already exist) to infer crossings. In this step, \textit{Pedestrianfer} iterates over each street intersection node in the street network and generates (1) all street lines associated with the intersection, directed outwards from the intersection and up to half of the distance to the next intersection, and (2) a set of candidate sidewalks to connect with a crossing on each side of each street. \textit{Pedestrianfer} then generates candidate crossings that are lines drawn from a sidewalk on the left side of a street to a sidewalk on the right side of a street. Multiple candidates are generated for each street of an intersection by selecting a series of points along that street (every 1 meter) and generating a crossing that connects the closest corresponding left and right sidewalks. Metrics of the candidates are generated, including the distance of the street point to the intersection, the crossing line length, and the angle between the crossing line and the street it crosses. A cost function is then used to heuristically select the \emph{best} crossing: one that minimizes a linear combination of the distance to the street intersection, crossing line length, and non-orthogonal crossing angles. Therefore, \textit{Pedestrianfer} estimates street crossing locations that are near intersections, which align with common (albeit not universal) policies and pedestrian safety measures. In the case that crossing locations with ground markings are already known and present in an associated dataset (as \textit{Point} data), \textit{Pedestrianfer} will generate a crossing near that point by projecting to the nearest known sidewalk candidates on each side.

Lastly, \textit{Pedestrianfer} splits crossing lines into three segments: (1) the originating sidewalk surface, (2) the street surface, and (3) the destination sidewalk surface. These correspond to typical surfaces a pedestrian would travel in an urban area. Sidewalk-street transitions, which often have vertical displacements or are where curb-cut ramps meet the street, can be interpreted as potential curb interfaces. \textit{Pedestrianfer} does not use any metadata to determine where to split each crossing into these 3 segments, but rather provides a low-information hypothesis on sidewalk-curb-street locations, which can be improved with manual mapping (GIS software, crowdsourcing in OSM) or learning-based method such as those described in Section \ref{sec:cnn} and Section \ref{sec:opt}.

\subsection{Segmentation Network}
\label{sec:cnn}
In order to verify, correct, and refine the \textit{Pedestrianfer} hypothesized path network graph, we use inference from a CNN-based segmentation network. The segmentation network has a siamese-like structure, allowing it to fuse and utilize the information from both the aerial satellite image and the street map tiles. As shown in Figure \ref{fig:flow_chart}, the segmentation network has two identical branches, the aerial satellite images are used as the input in one branch and the street map tiles are used in the other. Each branch has its encoder-decoder structure. In addition, to fuse the information from both branches, we concatenate feature maps from both branches at different layers hierarchically during the up-sample process. Experiments described in Section \ref{sec:exp} used FCN-8s \cite{long2015fully} as the backbone model, but it can be replaced by any other segmentation model with a similar structure.

\subsection{Optimization for Full Pedestrian Path Network graph}
\label{sec:opt}

\subsubsection{Optimizing node geolocation}
\textit{Pedestrianfer} generates a hypothesized path network graph that outlines the potential location and connectivity of sidewalks and crossings. To obtain a more accurate pedestrian path network, we use information from the segmentation network to refine the hypothesized path network graph. Our strategy is to first find the optimized geolocation of the nodes in the graph, then connect the nodes with edges to complete the graph.  

\begin{figure}[htbp] 
    \centering
        \begin{subfigure}[]{0.45\columnwidth}
        \centering
        \caption{}
        \includegraphics[width=\columnwidth]{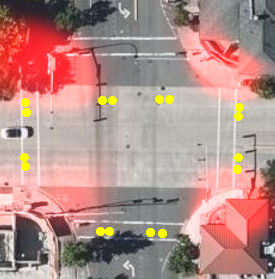}
        \label{fig:node_hypo}
    \end{subfigure}
    \begin{subfigure}[]{0.45\columnwidth}
        \centering
        \caption{}
        \includegraphics[width=\columnwidth]{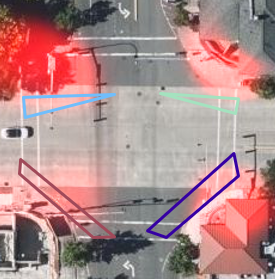}
        \label{fig:node_before}
    \end{subfigure}
    \begin{subfigure}[]{0.45\columnwidth}
        \centering
        \caption{}
        \includegraphics[width=\columnwidth]{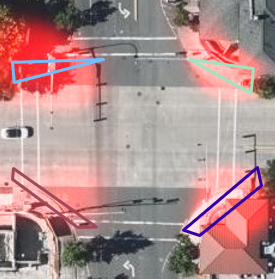}
        \label{fig:node_after}
    \end{subfigure}
    \begin{subfigure}[]{0.45\columnwidth}
        \centering
        \caption{}
        \includegraphics[width=\columnwidth,height=110px]{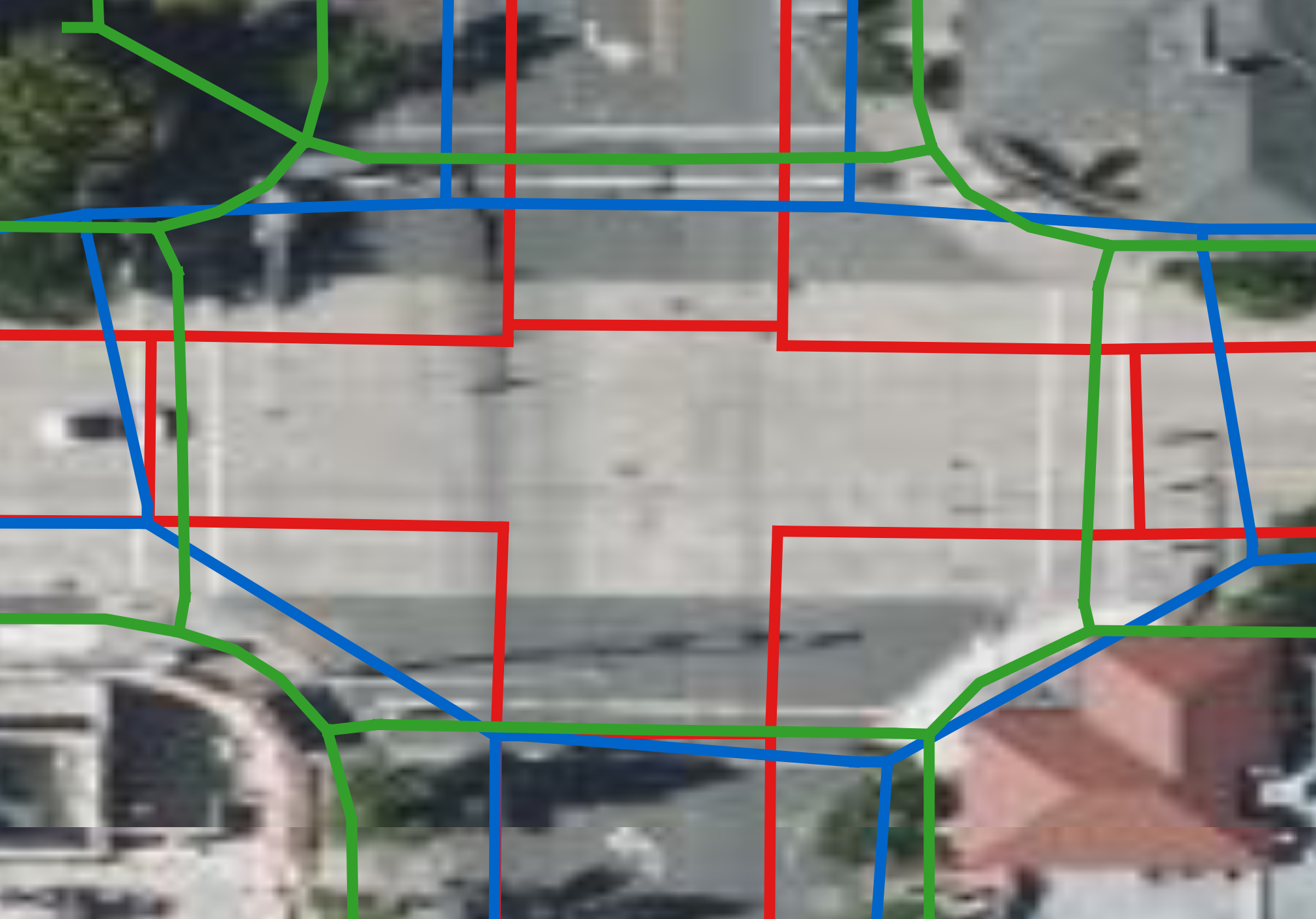}
        \label{fig:node_graph}
    \end{subfigure}
    
    \caption{Illustration of node location optimization, the probability of each pixel being the \textit{corner bulb} is shown as a heat map in red in (a) - (c). (a) \textit{Pedestrianfer} hypothesized nodes (b) Polygons formed by the hypothesized nodes in each corner  (c) New set of nodes optimized with information from the segmentation network (d) Compare to Human annotation: \colorbox{Green}{Green} is human annotation graph \colorbox{Red}{Red} is Pedestrianfer Hypothesized graph. \colorbox{Blue}{Blue} is the optimized graph.}
    \label{fig:sample_opt}
\end{figure}

At each intersection, there are nodes representing sidewalks endpoints and curbs in a hypothesized graph generated by \textit{Pedestrianfer}. Ideally in a correct pedestrian path network, these nodes should all locate in a connected region being segmented as the \textit{corner bulb} class (an example is given in Figure \ref{fig:sample_opt} (c) and  Figure \ref{fig:sample_opt} (d)). To infer the correct geolocation of these nodes, we aim to find a parameterized affine transformation to warp each set of hypothesized nodes in a corner, represented by their pixel coordinates, to a new set of coordinates, so they better align with the \textit{corner bulb} class in the predicted segmentation mask.  We start by connecting the nodes at each corner (shown in Figure \ref{fig:node_hypo}) to form a closed polygon (shown in Figure \ref{fig:node_before}), then we find a parameterized affine transformation so that the sum of the probability of each pixel under each closed polygon being the \textit{corner bulb} class is the greatest.

Mathematically, this optimization process can be defined as follows. For the total of $n$ points  $[(x_1, y_1), (x_2, y_2), ... (x_n, y_n)]$ in a corner in a given Image $I$, the affine transformation that warps them to a set of new points $[(x'_1, y'_1), (x'_2, y'_2), ... (x'_n, y'_n)]$ can be described as: 

\begin{equation}
\begin{bmatrix}
x'_1 & ... & x'_n\\
y'_1 & ... & y'_n
\end{bmatrix}
= 
\begin{bmatrix}
a & b\\
c & d
\end{bmatrix}
*
\begin{bmatrix}
x_1 & ... & x_n\\
y_1 & ... & y_n
\end{bmatrix}
+
\begin{bmatrix}
t_1\\
t_2
\end{bmatrix}
= AX + t
\end{equation}

In words, $X$ is a set of points representing the coordinates of the nodes in a hypothesized graph at a given street corner. The transformation parameters $A$ and $t$, and the new set of coordinates $[(x'_1, y'_1), (x'_2, y'_2), ... (x'_n, y'_n)]$ are to be found using the information obtained from the segmentation network.

Assuming there are a total of $m$ pixels that fall into the polygon enclosed by the $n$ corner points $X$, and the probability (as predicted by the segmentation network) that each of these $m$ image pixels to be the \textit{corner bulb} class is denoted as $p_i$, then we define the function that finds these pixels and their probability as $f$:

\begin{equation}
f: X  \longmapsto [p_1, p_2, ... p_m]
\end{equation}

and define a function $g$ that sums $[p_1, p_2, ... p_m]$ as:
\begin{equation}
g(f(AX + t)) = g([p_1, p_2, ... p_m]) = \sum ^m_{i=1} p_i 
\end{equation}

The goal is to maximize $g$ so that the pixels under the new polygons have the greatest sum of probability (shown in Figure \ref{fig:node_after}). Thus, the optimization problem can be expressed as: 

\begin{equation} \label{eq:opt}
\begin{split}
& \underset{A,t}{\text{minimize}}  \quad -g\left(f\left( AX + t \right)\right) \\
& \text{subject to} \quad \forall i \in [1,n] \quad 0 < x'_i < I_{width}, 0< y'_i < I_{height}
\end{split}
\end{equation}

There is no closed-form solution to Equation \ref{eq:opt}, and the gradient of $f$ cannot be explicitly found. Hence we use the simultaneous perturbation stochastic approximation (SPSA) \cite{spall1998implementation} method to find the optimal parameters $A$ and $t$ of the objective function $g$.

After optimizing sidewalk and curb node geolocation, we connect them with the information from the hypothesized graph to generate new sidewalks and crossing edges. The optimization process is shown in Figure \ref{fig:sample_opt}. Figure \ref{fig:node_hypo} and \ref{fig:node_before} show the original node locations in the \textit{Pedestrianfer} hypothesized graph. In this example, \textit{Pedestrianfer} approximates the nodes erroneously in the middle of the road. Figure \ref{fig:node_after} shows the locations of the nodes post optimization, where the nodes and the enclosed polygon formed by the nodes are moved to the middle of a region predicted as the \textit{corner bulb} class. The improved graph (post-optimization) in Figure \ref{fig:node_graph}, shows graph elements considerably closer to the human-generated (ground truth) graph. A more detailed quantitative evaluation is made in Section \ref{sec:eval_graph}. 

\subsubsection{Graph refinement with class probability}
\label{sec:opt_node_loc}
The optimized graph is used with probability masks of each class from the segmentation network to further improve the predicted path network map accuracy. 

For a set of hypothesized nodes at a given corner, if the closed polygon they form does not overlap with enough high-probability pixels representing the \textit{corner bulb class}, we consider these nodes to be falsely hypothesized. Mathematically, we define $\mu_{p}$ as:
\begin{equation}
\mu_{p}= g(p)/m
\label{eq:avg_prob}
\end{equation}
If $\mu_{p}$ is less than a set threshold, these nodes are considered to be false-hypothesized and therefore directly removed from the graph without optimizing with equation \ref{eq:opt}. The threshold is chosen to balance the edges' precision and recall in this post-processing step. Higher thresholds lead to higher precision. 

For downstream applications (e.g. wayfinding) that use the predicted graph, confidence values are added to edges to inform the optimization for high-confidence routes, i.e., the application may choose to avoid low-confidence edges and instead optimize for higher-confidence paths. For each edge created by connecting two nodes in the graph, we assign a confidence value as an attribute of the edge as follows: each \textit{LineString} representing the \textit{crossing} class, is converted into a polygon by adding a buffer to each side of the \textit{LineString}, then we compute the mean probability of the pixels in the polygon in the \textit{crossing} class, similar to Equation \ref{eq:avg_prob}. The mean probability is used as the confidence value and stored as an attribute of the edge in the graph. Similarly, the confidence values of the \textit{sidewalk} edges are computed and stored as an attribute of the \textit{sidewalk} edges. The confidence value improves the graph for various transportation network analyses, sidewalk network scoring, personalized routing \cite{bolten2019accessmap}, and other downstream applications.

\section{Experiments}
\label{sec:exp}

\begin{figure*}[htbp]
    \centering
    \resizebox{2\columnwidth}{!}{
    \begin{tabular}{c|c|c|c|c|c}
     \textbf{Aerial satellite image} & \textbf{Street map image} & \textbf{Ground Truth}  & \textbf{Satellite only} & \textbf{Street only} & \textbf{Satellite + Street}\\
        \includegraphics[width=0.33\columnwidth]{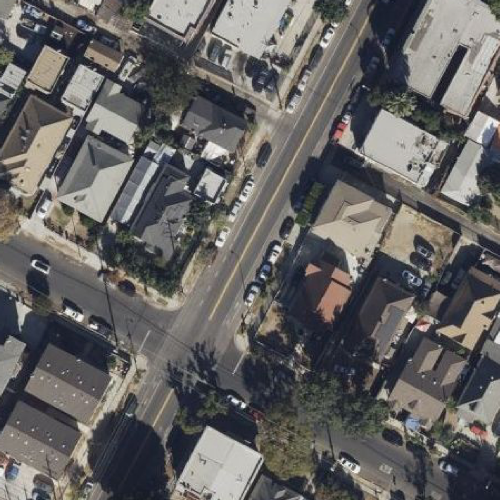}   &
        \includegraphics[width=0.33\columnwidth]{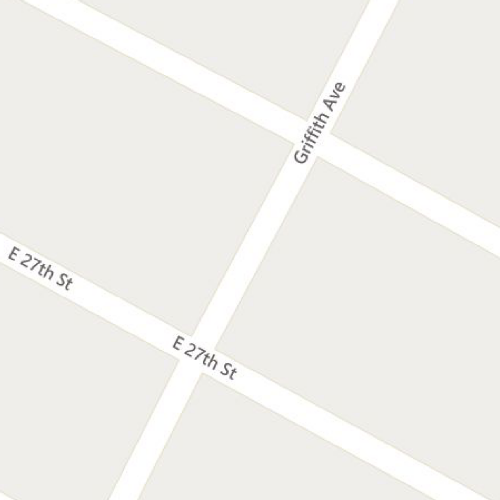} &
         \includegraphics[width=0.33\columnwidth]{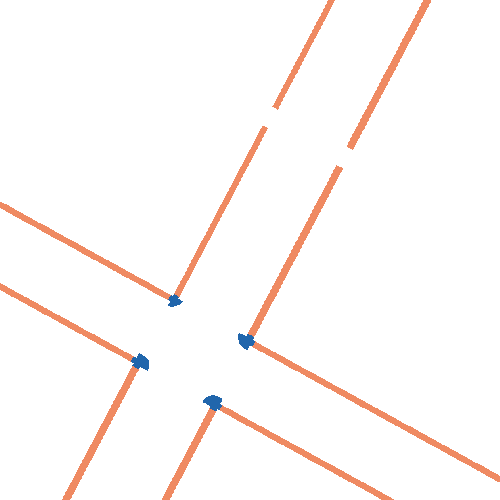} &
        \includegraphics[width=0.33\columnwidth]{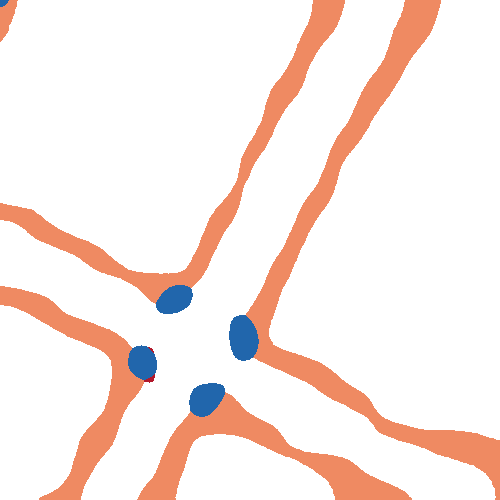} &
        \includegraphics[width=0.33\columnwidth]{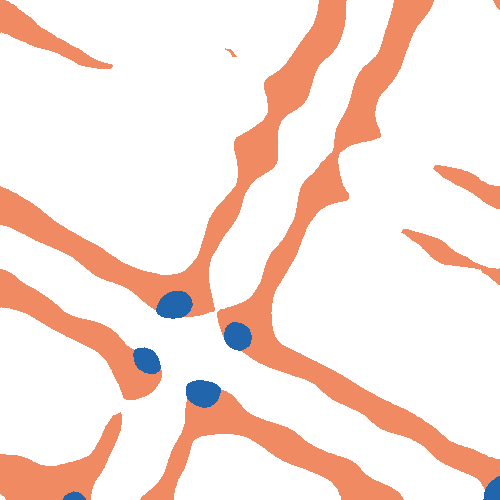} &
        \includegraphics[width=0.33\columnwidth]{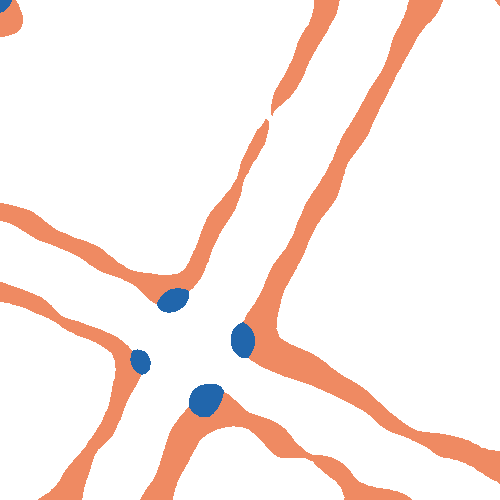} \\
        \includegraphics[width=0.33\columnwidth]{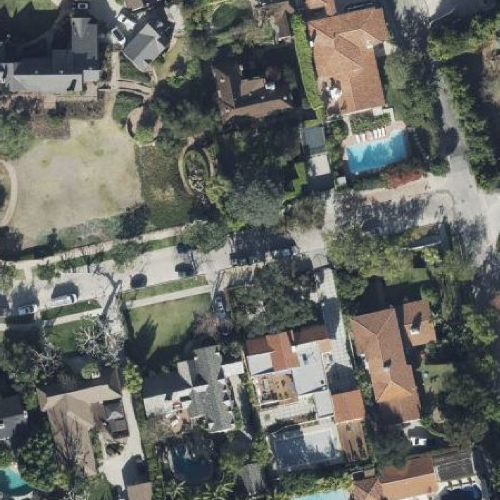}   &
        \includegraphics[width=0.33\columnwidth]{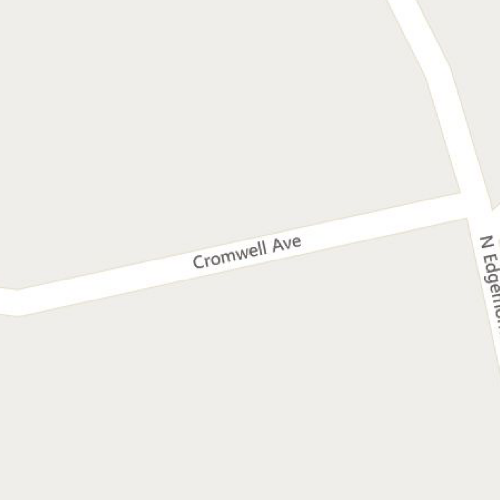} &
         \includegraphics[width=0.33\columnwidth]{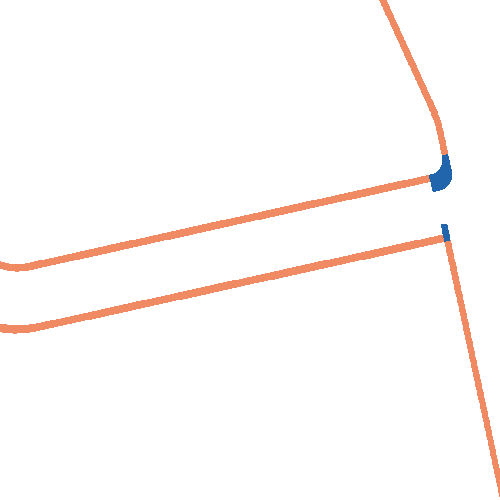} &
        \includegraphics[width=0.33\columnwidth]{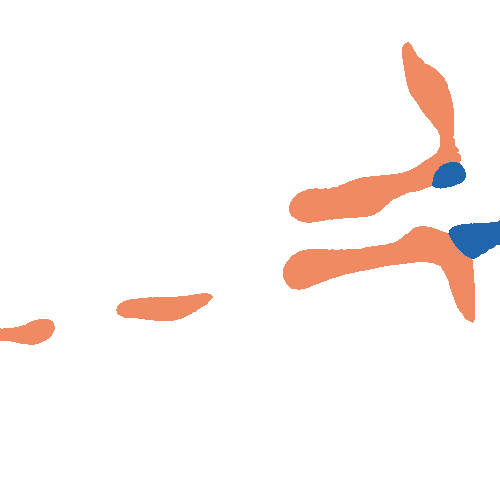} &
        \includegraphics[width=0.33\columnwidth]{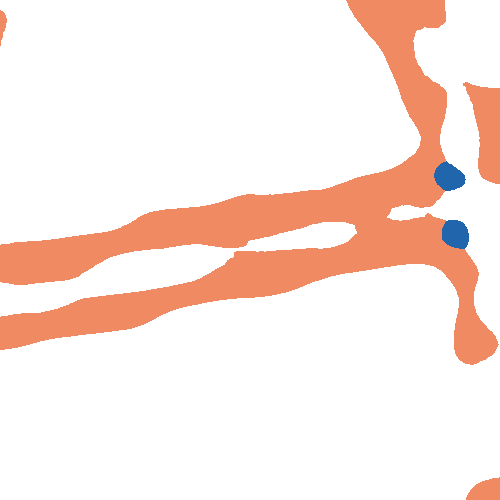} &
        \includegraphics[width=0.33\columnwidth]{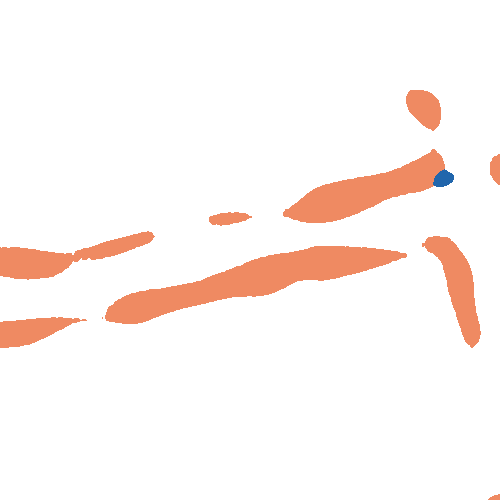} \\
        \includegraphics[width=0.33\columnwidth]{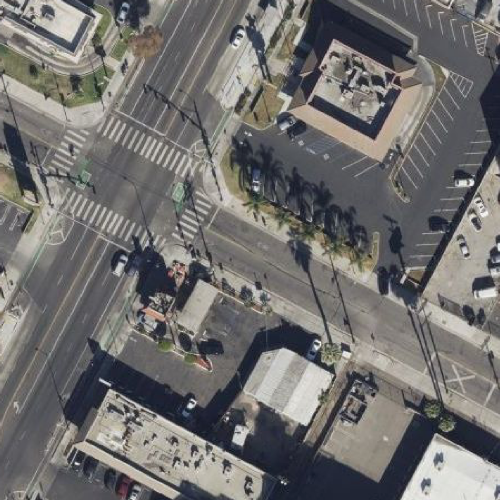}   &
        \includegraphics[width=0.33\columnwidth]{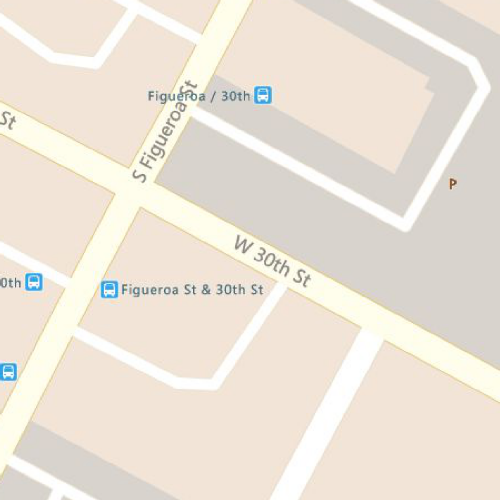} &
         \includegraphics[width=0.33\columnwidth]{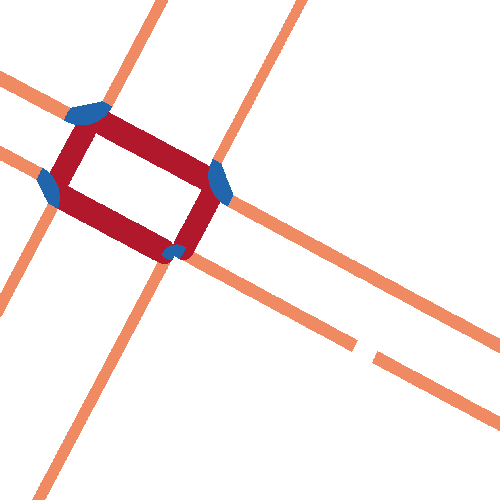} &
        \includegraphics[width=0.33\columnwidth]{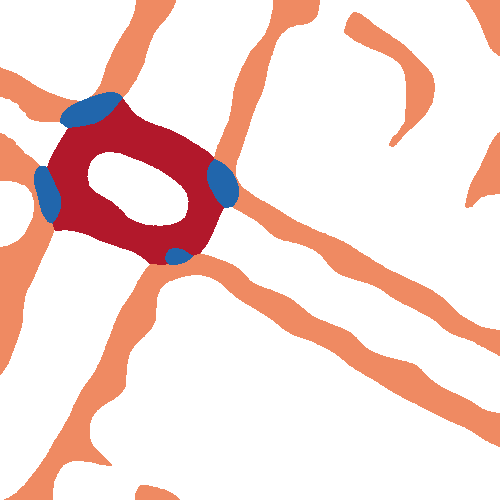} &
        \includegraphics[width=0.33\columnwidth]{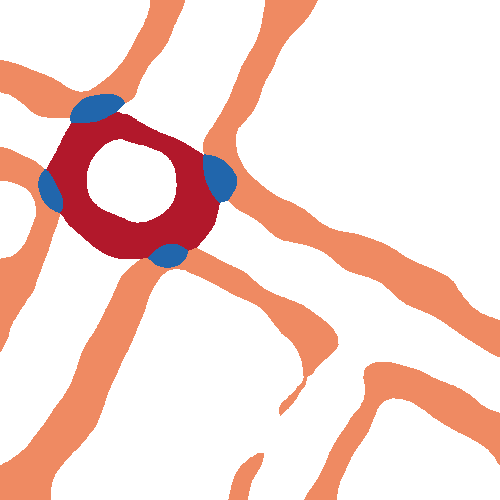} &
        \includegraphics[width=0.33\columnwidth]{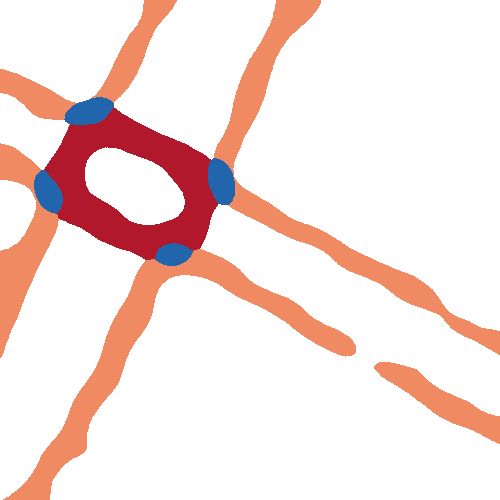} \\
        \includegraphics[width=0.33\columnwidth]{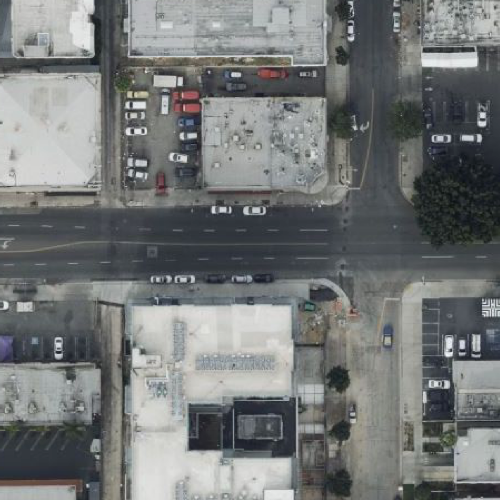}   &
        \includegraphics[width=0.33\columnwidth]{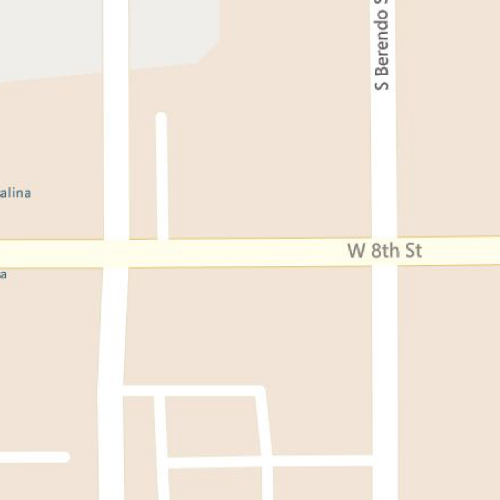} &
         \includegraphics[width=0.33\columnwidth]{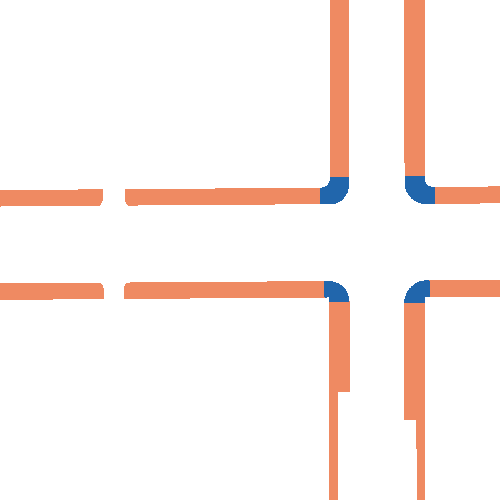} &
        \includegraphics[width=0.33\columnwidth]{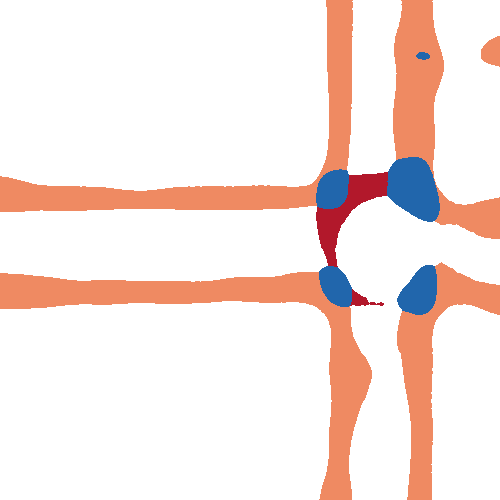} &
        \includegraphics[width=0.33\columnwidth]{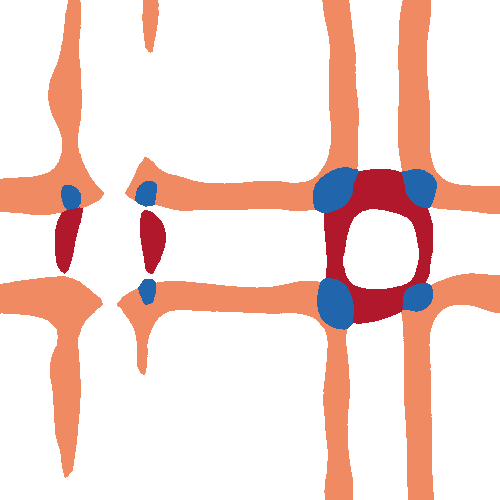} &
        \includegraphics[width=0.33\columnwidth]{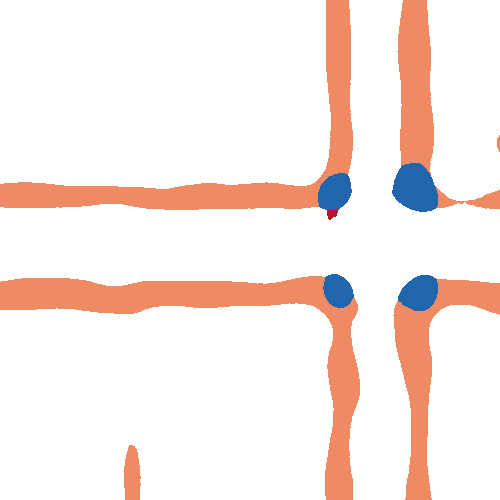} \\
        % removing the fifth example (which is simillar to the fourth, running out ouf space)
        % \includegraphics[width=0.33\columnwidth]{images/qualitative/36_sat.png}   &
        % \includegraphics[width=0.33\columnwidth]{images/qualitative/36_st.png} &
        %  \includegraphics[width=0.33\columnwidth]{images/qualitative/36_gt.png} &
        % \includegraphics[width=0.33\columnwidth]{images/qualitative/36_prediction_sat.png} &
        % \includegraphics[width=0.33\columnwidth]{images/qualitative/36_prediction_street.png} &
        % \includegraphics[width=0.33\columnwidth]{images/qualitative/36_prediction_both.png} \\
 \end{tabular}
 }
    \caption{Qualitative results on the validation set. The segmentation results of 3 different models are shown in columns 4-6. (1) Trained with the aerial satellite image branch only (2) Trained with the street map tile branch only (3) Trained with both the aerial satellite image branch and the street map image tile branch. The model that uses both aerial satellite images and street map images generates better predictions than the models that use only one input. A detailed discussion of these samples is made in Section \ref{sec:model_qual}. Quantitative analysis of models' performance is made in Section \ref{sec:model_quan} and Table \ref{tab:quan}. }
    \label{fig:qual_samples}
\end{figure*}

 \begin{table*}
    \centering
    \caption{Quantitative segmentation results: the model that uses both aerial satellite images and street map tiles outperforms models that use only one branch of data}
    \label{tab:quan}
\centering
    \begin{tabular}{l|cccccc|cc}
    \toprule
       \multirow{2}{*}{\textbf{Method}} &\multicolumn{6}{c|}{\textbf{Pixel-wise}} &\multicolumn{2}{c}{\textbf{Instance-wise}} \\ 
       &  \textbf{BG IoU} & \textbf{SW IoU} & \textbf{Corner IoU} & \textbf{Crossing IoU} & \textbf{mIoU} & \textbf{Accuracy} & \textbf{Corner precision} & \textbf{Corner recall}\\
       \midrule
       \textbf{Satellite Only} & 0.89 & 0.42 & 0.47 & 0.42 & 0.55 & 0.89 & 0.59 & 0.68\\
        \textbf{Street Only} & 0.85 & 0.33 & 0.48 & 0.37 & 0.51 & 0.86 & 0.60 & 0.63\\
        \textbf{Satellite + Street} & 0.91 & 0.47 & 0.57 & 0.49 & \textbf{0.61} & 0.92 & 0.73 & 0.69\\
    \bottomrule
\end{tabular}
\end{table*}

\begin{table*}
\centering
\caption{Quantitative evaluation: graph level analysis}
\label{tab:graph_quan}
\begin{subtable}[b]{2\columnwidth}
    \centering
    \caption{Sidewalks}
    \begin{tabular}{l|ccc|ccc|ccc}
    \toprule
       & \multicolumn{3}{c|}{Precision} & \multicolumn{3}{c|}{Recall} & \multicolumn{3}{c}{F1}\\
         & \textbf{LA} & \textbf{Bellevue} & \textbf{Quito} & \textbf{LA} & \textbf{Bellevue} & \textbf{Quito} & \textbf{LA} & \textbf{Bellevue} & \textbf{Quito}\\
       \midrule
       \textbf{Pedestrianfer (Baseline)} & 68.1 & 69.2 & 67.1 & 73.4 & 73.2 & 71.2 & 70.7 & 71.1 & 69.1 \\
        \textbf{Pedestrianfer + Segmentation } & 83.5 & 82.4 & 75.6 & 77.8 & 77.0 & 73.6 & 80.5 & 79.6 & 74.6 \\
    \bottomrule
\end{tabular}
\label{tab:graph_quan_sw}
\end{subtable}

\begin{subtable}[b]{2\columnwidth}
    \centering
    \caption{Crossings}
    \begin{tabular}{l|ccc|ccc|ccc}
    \toprule
       & \multicolumn{3}{c|}{Precision} & \multicolumn{3}{c|}{Recall} & \multicolumn{3}{c}{F1}\\
         & \textbf{LA} & \textbf{Bellevue} & \textbf{Quito} & \textbf{LA} & \textbf{Bellevue} & \textbf{Quito} & \textbf{LA} & \textbf{Bellevue} & \textbf{Quito}\\
       \midrule
       \textbf{Pedestrianfer (Baseline)} & 73.5 & 69.2 & 71.9 & 82.1 & 71.8 & 75.8 & 77.6 & 70.5 & 73.8 \\
        \textbf{Pedestrianfer + Segmentation } & 76.7 & 72.2 & 74.4 & 87.7 & 75.1 & 78.7 & 81.8 & 73.6 & 76.5 \\
    \bottomrule
\end{tabular}
\label{tab:graph_quan_crossing}
\end{subtable}

\end{table*}

\subsection{Training the Segmentation Network}
\label{sec:exp_train}
Three segmentation networks were trained with different setups:
aerial satellite image branch only, street map image tile branch only, and both aerial and street map image tile branches. All three models used the same dataset split and data augmentation techniques. Performance comparisons are shown in Figure \ref{fig:qual_samples} and Table \ref{tab:quan}.

\subsubsection{Dataset}
\label{sec:exp_data}
We train and validate with an $80\%/20\%$ split of the APE imagery and annotation samples in the Los Angeles area. Specifically, we chose to use the rasterized data from the City of Los Angeles annotations because of the amount of data available. Bellevue and Quito annotations from OpenSidewalks are used as unseen test sets.

\subsubsection{Data Augmentation}
\label{sec:exp_aug}
To improve the robustness of the segmentation network, data augmentations were applied to the training data during the training stage. First, random rotating, cropping, and resizing were applied. Again, we note the resolution differential in open aerial satellite images in North America (Los Angeles and Bellevue) as compared to those from South America (Examples are shown in Figure \ref{fig:data_sample}). 
We enhanced the segmentation network's performance on low-resolution images by using Gaussian kernels with random sizes to artificially generate low-resolution images in training.

\subsection{Qualitative Analysis}
\label{sec:model_qual}
Figure \ref{fig:qual_samples} visualizes the segmentation results on the validation set. The models' shape segmentation and identified locations for \textit{sidewalk}, \textit{crossing}, and \textit{corner bulb} align well with the ground-truth segmentation. These qualitative examples show the difficulty in predicting pedestrian path network classes with single-source input and the improvement gained from adding other input sources.
The first and second rows of Figure \ref{fig:qual_samples} show that different kinds of errors are produced when the model is trained with only one branch of input data. In the example shown in the first row, the prediction made with the model trained with only street images generated false-positive \textit{sidewalk} predictions. Adding aerial satellite images in model training helped remove these spurious predictions. Predictions made with only aerial satellite images incurred other limitations, as shown in the second-row example, when sidewalks were occluded in the satellite image by vegetation, the model trained with only those images generated false-negative \textit{sidewalk} predictions. Adding street image tiles in training helped predict many of the occluded sidewalks. To further demonstrate the effectiveness of using both the aerial satellite images and the street image tiles as inputs to the model, the examples in row 3 and row 4 of Figure \ref{fig:qual_samples} show different cases in which using both branches of data generated better result than using one branch of input data alone. In the third row example, the model trained with only one branch of data generated different false-positive \textit{sidewalk} predictions, and using both branches of data to train the model helped remove these false-positive predictions. Similarly, in the fourth-row example, models trained with only one branch of data generated false-positive predictions on \textit{sidewalk} and \textit{crossing}, while using both branches of data helped remove these false predictions. A quantitative analysis of the performance of our method when testing in different cities is provided in Section \ref{sec:eval_graph}.
% While the segmentation outcome is best on the validation set.  the model can still well segment the \textit{sidewalk}, \textit{crossing}, and \textit{corner bulb} when tested on the unseen datasets where the images contain different built structures and have lower resolutions as shown in Figure \ref{fig:qual_samples_unseen}. 

\subsection{Quantitative Analysis}
\label{sec:model_quan}
Table \ref{tab:quan} shows the quantitative experiment results for the models trained with the 3 different setups: (1) trained with the aerial satellite image branch only, (2) trained with the street map image tile branch only, and (3) trained with both the aerial satellite images and street map image tiles. The metrics we included are (1) overall mIoU, (2) the mIoU for each of the 4 classes (\textit{background} (BG), \textit{sidewalk} (SW), \textit{corner bulb}, and, \textit{crossing}) in the dataset, (3) pixel accuracy, and (4) instance-wise precision and recall for the important \textit{corner bulb} class. 

From Table \ref{tab:quan} we observe that the model trained with both the aerial satellite images and street map image tiles has a higher mIoU on each of the classes than the models trained with only one branch of data, demonstrating its better pixel-wise performance. In addition, higher precision and recall on the important \textit{corner bulb} class demonstrates its better instance-wise performance. These metrics combined shows the effectiveness of the model for segmenting the pedestrian environment with both the aerial satellite images and the street map image tiles, confirming the discussions we made in Section \ref{sec:model_qual} based on the observations from Figure \ref{fig:qual_samples}. A graph-level analysis is made in Section \ref{sec:eval_graph}.

\subsection{Quantitative Evaluation of the Inferred Path Network Graph}
\label{sec:eval_graph}
Although the models perform well with the pixel-wise and instance-wise measures, these metrics do not fully reflect the accuracy of the connected graph prediction. mIoU (or other pixel-wise measures) cannot measure how close a predicted graph is to the ground truth graph. In order to measure the similarity between the predicted graphs to the ground truth graph, we compare (1) the pedestrian path network graph generated by \textit{Pedestrianfer}, and (2) the graph optimized using the segmentation network, to the ground truth graph generated and validated by human mappers from the OpenSidewalks project. \textit{Pedestrianfer} is used as the baseline method as it is similar to the method proposed by \cite{li2018semi} and it represents the methods that derive a pedestrian path network purely from existing street network information. The metrics we used for evaluating the graphs are similar to the ones proposed by \cite{mattyus2017deeproadmapper} for road topology measurements, namely the precision, recall, and F1 score based on the assignments of predicted sidewalk edges (or crossing edges) to the corresponding edges in the ground truth graph. The results are summarized in Table \ref{tab:graph_quan}. The first row of each sub-table shows the performance of the baseline model, and the second row of each sub-table shows the performance of our method where the information from the segmentation networks that use both aerial satellite images and street images is used to optimize the predicted graphs. Table \ref{tab:graph_quan_sw} shows evaluation results on sidewalks and Table \ref{tab:graph_quan_crossing} shows evaluation results on crossings. The graph generated with the segmentation network outperforms the baseline model, especially in the precision category because the baseline model that uses only street network information tends to be overly optimistic about the existence of pedestrian paths. High precision, recall, and F1 score demonstrate the effectiveness of our method in inferring a pedestrian path network with both the existing street network and the information learned with the segmentation network. In addition, our method maintains high precision and recall when tested on unseen datasets (Bellevue and Quito) and outperforms the baseline methods, despite the test areas having significantly different built environments and aerial satellite image resolution (with notably low-resolution imagery in Quito). Most importantly, since we start from a connected graph hypothesized by \textit{Pedestrianfer} and utilize the segmentation network alone to correct node locations and remove false-hypothesized geometries, the outcome graph stays as a connected graph that can be directly used as a routable network graph.

% We measure the correctness of the graph by matching each of the sidewalk edges (or crossing edges) to the corresponding edge in the ground truth graph. Then interpolate N equal-distance points on the edge LineString, and compute the avg distance of the N interpolated point to ground truth edge LineString.  

\section{Conclusion}
\label{sec:dis}
In this work, we introduce the APE dataset to address the dearth of data and methods to automate mapping for pedestrian path networks. APE includes aerial images, street map tiles, and annotated pedestrian environments. It provides the research community with a new and challenging dataset to address pedestrian network path predictions through machine learning methods. The dataset covers diverse urban areas and we demonstrated it can be used for computer vision tasks. Additionally, we developed a method to infer pedestrian path networks using segmentation and street data, and validated the method's accuracy through comparison to human-annotated data. Through our experiments, we were able to both metricize human performance in pedestrian networks annotation tasks like drawing geometries and labeling
(metrics achieved through double annotations) and also evaluated the baseline performance of a commonly used segmentation model. The APE dataset and our process provide valuable initial contributions to pedestrian environment research, specifically in contributing to the many wayfinding and planning tasks that rely on detailed pedestrian transportation networks. We hope this work will inspire further pedestrian network data collections at scale.

%%
%% 
% *** Ricky: Can we remove acknowledgments for now because we don't have space. We should try make some space and add it back, or find a way to add it back in the camera-ready version. ***
\begin{acks}
This work was funded in part by the Taskar Center for Accessible Technology, USDOT ITS4US NOFO No: 693JJ322NF00001, and Microsoft's AI4Accessibility award. Thanks to the G3ict organization for its support of the Los Angeles, Quito, Sao Paulo, Santiago, and Gran Valparaiso, on-the-ground mapping efforts.
\end{acks}

\bibliographystyle{ACM-Reference-Format}
\bibliography{main}

\end{document}